%% file: main.tex
\documentclass[10pt,twocolumn,letterpaper]{article}

\pdfoutput=1

\usepackage[pagenumbers]{cvpr} %

\input{preamble}

\definecolor{cvprblue}{rgb}{0.21,0.49,0.74}
\usepackage[pagebackref,breaklinks,colorlinks,citecolor=cvprblue]{hyperref}

\newcommand\blfootnote[1]{%
\begingroup 
\renewcommand\thefootnote{}\footnote{#1}%
\addtocounter{footnote}{-1}%
\endgroup 
}

\usepackage[capitalize]{cleveref}
\crefname{section}{Sec.}{Secs.}
\Crefname{section}{Section}{Sections}
\Crefname{table}{Table}{Tables}
\crefname{table}{Tab.}{Tabs.}
\Crefname{figure}{Figure}{Figures}
\crefname{figure}{Fig.}{Figs.}

\def\paperID{4584} %
\def\confName{CVPR}
\def\confYear{2024}

\title{\algoNameFull: Spatially Aware Multi-View Diffusers}

\author{
Yash Kant$^{1,2,4}$, Ziyi Wu$^{1,4}$, Michael Vasilkovsky$^2$, Guocheng Qian$^{2,3}$, Jian Ren$^2$, \\
Riza Alp Guler$^2$, Bernard Ghanem$^3$, Sergey Tulyakov$^{2,*}$, Igor Gilitschenski$^{1,4,*}$, Aliaksandr Siarohin$^{2,*}$ \\
$^1$University of Toronto, $^2$Snap Research, $^3$ KAUST, $^4$Vector Institute \\
\url{https://yashkant.github.io/spad/} \\
}

\begin{document}

\twocolumn[{%
    \maketitle
    \vspace{-6mm}
    \begin{figure}[H]
        \hsize=\textwidth
        \centering
        \includegraphics[width=2\linewidth]{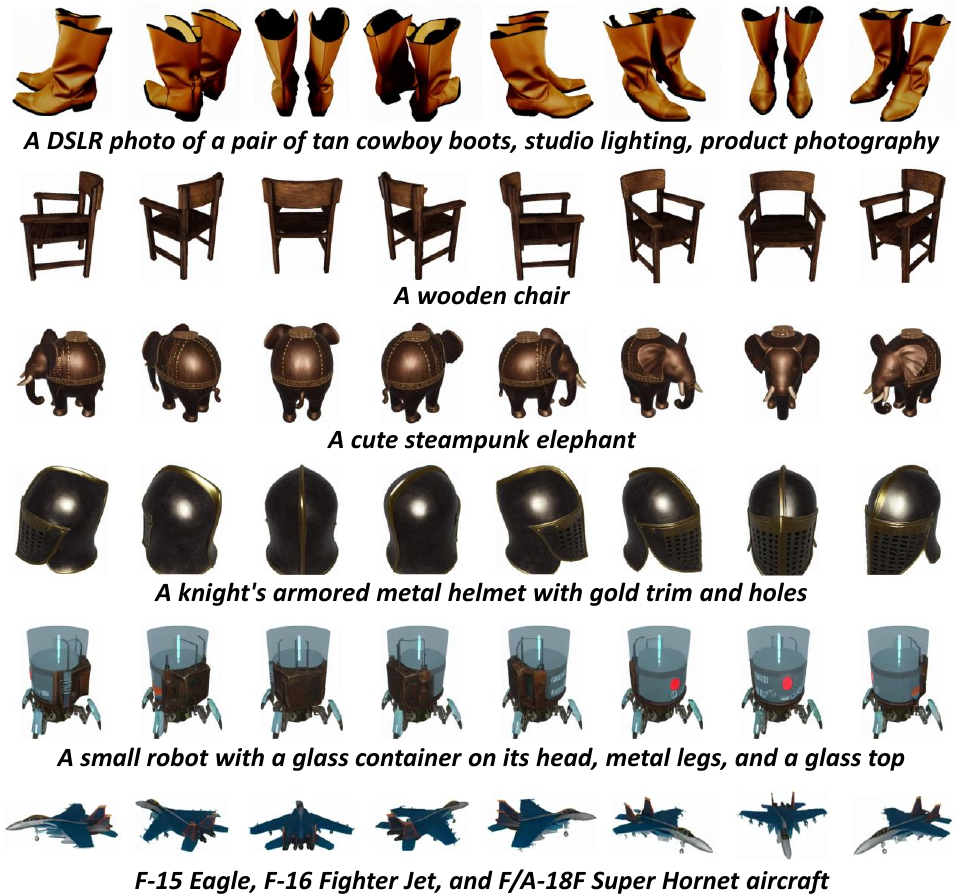}
        \vspace{-1mm}
        \caption{
            \textbf{Consistent multi-view generation from text with \algoNameFull.}
            Given a text prompt, \algoNameFull is capable of synthesizing many 3D consistent images of the same object, ranging from daily objects to highly complex machines. \algoNameFull can generate many images from arbitrary camera viewpoints, while being trained only on four views. Here, we generate eight views sampled uniformly at a fixed elevation.
        }
        \label{fig:teaser}
    \end{figure}
}]

\maketitle

\input{sections/0_abstract}

\input{sections/1_intro}

\input{sections/2_related}

\input{sections/3_method}

\input{sections/4_experiment}

\input{sections/5_conclusion}

\subsection*{Acknowledgments}

\noindent We would like to thank Xuanchi Ren and Weize Chen, for valuable discussions and support.

{
    \small
    \bibliographystyle{ieeenat_fullname}
    \bibliography{refs}
}

\clearpage
\input{sections/6_appendix}

\end{document}

%% file: preamble.tex
\usepackage{graphicx}
\usepackage{amsmath}
\usepackage{amssymb}
\usepackage{booktabs}
\usepackage{float}
\usepackage{enumerate}

\usepackage{mathtools} %
\usepackage{booktabs} %
\usepackage{tikz} %
\usepackage{amsfonts}
\usepackage{multirow}
\usepackage{multicol}
\usepackage{color,colortbl}
\usepackage{xcolor}
\usepackage{xspace}
\usepackage{wrapfig}
\usepackage{subcaption}
\usepackage{bbm}
\usepackage{bm}
\usepackage{url}

\newcommand{\algoNameFull}{SPAD\xspace}

\def\eg{e.g.,\ }               %
\def\ie{i.e.,\ }               %

\definecolor{demphcolor}{RGB}{100,100,100}
\newcommand{\demph}[1]{\textcolor{demphcolor}{#1}}
\newcommand{\pms}[2]{#1{\footnotesize{{\demph{{$\pm$#2}}}}}}

\def\cam{\bm{E}}
\def\deltacam{\Delta \cam}
\def\ray{\bm{r}}
\def\camorigin{\bm{o}}
\def\camdir{\bm{d}}
\def\pluemb{\bm{P}}

\def\img{\bm{x}}
\def\noise{\bm{\epsilon}}
\def\textemb{\bm{y}}
\def\srcpoint{s}
\def\epiline{l}
\def\featmap{\bm{F}}
\def\dmmodel{\epsilon_\theta}

\def\degree{^{\circ}}

\usepackage{soul}

\newcommand{\heading}[1]{\noindent\textbf{#1}}

\newlength\pagetopmargin
\newlength\figcapmargin
\newlength\figmargin
\newlength\tablecapmargin
\newlength\tablemargin

\newcolumntype{H}{>{\setbox0=\hbox\bgroup}c<{\egroup}@{}}

%% file: sections/0_abstract.tex
\begin{abstract}
We present \algoNameFull, a novel approach for creating consistent multi-view images from text prompts or single images. 
 To enable multi-view generation, we repurpose a pretrained 2D diffusion model by extending its self-attention layers with cross-view interactions, and fine-tune it on a high quality subset of Objaverse.
We find that a naive extension of the self-attention proposed in prior work (\eg MVDream~\cite{MVDream}) leads to content copying between views. 
Therefore, we explicitly constrain the cross-view attention based on epipolar geometry. 
To further enhance 3D consistency, we utilize Plücker coordinates derived from camera rays and inject them as positional encoding. This enables \algoNameFull to reason over spatial proximity in 3D well.
In contrast to recent works that can only generate views at fixed azimuth and elevation, \algoNameFull offers full camera control and achieves state-of-the-art results in novel view synthesis on unseen objects from the Objaverse and Google Scanned Objects datasets.
Finally, we demonstrate that text-to-3D generation using \algoNameFull prevents the multi-face Janus issue.
\blfootnote{* Equal supervision}

\end{abstract}

%% file: sections/1_intro.tex
\section{Introduction}

3D content generation holds great importance in a wide range of applications, including gaming, virtual reality, manufacturing, \etc.
Yet, the creation of high-quality 3D assets remains a time-consuming endeavor even for seasoned 3D artists. 
In recent years we have witnessed the emergence of generative models capable of creating 3D objects from a single or several 2D images, or just text inputs.
Early methods in this field directly train models such as Variational Auto Encoders (VAEs)~\cite{VAE} and Generative Adversarial Networks (GANs)~\cite{GANs} on 3D shapes~\cite{TrainOn3D1,TrainOn3D2,OccupancyNetworks,PointCloudAE}. 
These methods produce results with lower resolution to manage computational demands and have limited diversity due to the small scale of training dataset. 
Later approaches explored differentiable rendering to learn 3D GANs from monocular images~\cite{HoloGAN,Giraffe,EG3D,piGAN,GET3D,DIB-R,EpiGRAF,3dgp}. 
These methods improved resolution, but only show impressive results on relatively few categories (\eg ShapeNet~\cite{ShapeNet} furniture).  

Recent advances in Diffusion Models (DMs) have revolutionized the field of 2D image generation~\cite{DDPM,adm,ImprovedDDPM}.
Trained on billions of image-text data, state-of-the-art models~\cite{stable_diffusion,imagen} learn generic object priors that enable high-quality and generalizable text-guided image generation.
Recent works thus seek to leverage such 2D priors to generate 3D objects.
One line of research proposes to optimize a NeRF~\cite{NeRF} model by distilling a pre-trained text-to-image DM via Score Distillation Sampling (SDS)~\cite{dreamfusion,SJC}, which enables single-view 3D reconstruction~\cite{RealFusion,Nerdi,Magic123} and direct text-to-3D synthesis~\cite{ProlificDreamer,magic3d,DreamTime-SDS-BetterOpt}.
However, these methods lack understanding of the underlying object structures.
The 2D prior provided by pre-trained DMs only considers one view at each optimization step, ignoring the geometric relationship across views.
Even with hand-crafted prompts specifying explicit viewpoints~\cite{dreamfusion},these methods continue to exhibit 3D inconsistencies, exemplified by issues such as the multi-faced Janus problem.

One natural solution is to equip 2D diffusion models with some form of 3D understanding.
Recent work Zero-1-to-3~\cite{zero1to3} proposes to condition Stable Diffusion~\cite{stable_diffusion} with one view and generate another one given the relative camera pose. 
However, conditioning in Zero-1-to-3 is performed by simply concatenating the input view, while disregarding any geometric priors.
An alternate approach based on depth warping was proposed in iNVS~\cite{iNVS}. 
It shows that, provided with an accurate depth map, one can establish dense correspondences between two views. 
This allows DMs to reconstruct high-quality novel views. 
Unfortunately, the generation quality of iNVS heavily relies on the precision of depth maps, while monocular depth estimation in itself is an unsolved problem.

Recent works~\cite{tang2023dift,zhang2023tale} have observed that Stable Diffusion can be utilized to obtain accurate image correspondences.
Self-attention layers of text-to-image DMs can be directly used for establishing correspondences \emph{within} the same image~\cite{Tumanyan_2023_CVPR}.   An interesting question to consider is whether the same layers can also find correspondences \emph{between} different views, which can enable 3D geometric understanding. 
For this, we can modify the original self-attention into multi-view self-attention by running it over the concatenated feature maps across views.
This approach trained on orthogonal multi-view images with known camera parameters can generate multiple novel views of the same object simultaneously, as shown in previous works such as MVDream~\cite{MVDream}.
However, we find such a model lacks precise camera control across views, and cannot generate arbitrarily free novel views. 
When tasked to generate two views that are close to each other (with significant overlap), such a model suffers from content copying problem -- where the content of one view is just copied from another view without modification (see~\cref{fig:ablation-qual}).

Inspired by \cite{EpipolarFeatureTransformer}, we design an Epipolar Attention layer, where feature map positions in one view can only attend to positions along the epipolar lines in other views.
By restricting the cross-view attention maps, these layers enable better camera control and produce different views at viewpoints close to each other.
While Epipolar Attention alone significantly improves 3D consistency, since epipolar lines do not have a direction, it still remains difficult for this model to disambiguate the direction of the camera ray.
This ambiguity leads to flipping in predicted views, as observed in iNVS~\cite{iNVS} which also used epipolar lines.
Motivated by recent works in the Light Field Networks~\cite{r2l,MobileR2L}, we propose to represent rays passing through each pixel in Plücker coordinates, which assigns unique homogeneous coordinates for each ray. 
We use these coordinates as positional embeddings inside Epipolar Attention layers. 
These embeddings for rays hitting opposite sides of the object provide a high negative bias for self-attention, essentially preventing it from utilizing information from the wrong side.
Additionally, Plücker embeddings also encourages pixels whose rays are close to each other to have similar representations, thus promoting self-attention to pick features from nearby positions.

Our method can operate in two modes: text-conditioned and image-conditioned.
In text-conditioned mode, \algoNameFull can simultaneously denoise several views given a text prompt. 
While in image-conditioned mode, given an image, \algoNameFull denoises several other views. 
In both cases, the architecture of our method stays the same, and only the input and output changes.
We evaluate \algoNameFull in the task of text-conditioned multi-view generation and image-conditioned novel view synthesis on Google Scanned Objects (GSO)~\cite{downs2022google} and an unseen 
subset of Objaverse~\cite{reizenstein2021common} datasets.
The results show that \algoNameFull is able to synthesize high-quality and 3D consistent images of objects.
Finally, we enable high-quality text-to-3D generation using \algoNameFull via a) feed-forward multi-view to 3D triplane generator, and b) multi-view Score Distillation Sampling similar to~\cite{MVDream}.

%% file: sections/2_related.tex
\section{Related Works}

\heading{3D Generative Models.}
3D generative modeling is a long-standing problem in computer vision and graphics.
Earlier works directly train generative models such as Variational Auto Encoders (VAEs)~\cite{VAE} on ground-truth 3D shapes~\cite{TrainOn3D1,TrainOn3D2,OccupancyNetworks,TrainOn3D4,TrainOn3D5,PointCloudAE}.
However, due to the small scale of 3D shape datasets, these methods produce less realistic and diverse results compared to their 2D counterparts.
With the rapid development of Generative Adversarial Networks (GANs)~\cite{GANs} and differentiable rendering, later works focus on learning 3D GANs from monocular images, showing impressive generation of multi-view images~\cite{HoloGAN,BlockGAN}, radiance fields~\cite{Giraffe,GRAM,EG3D,MVCGAN,VolumeGAN,StyleNeRF,piGAN}, and meshes~\cite{ConvMesh,Textured3DGAN,GET3D,DIB-R,DIB-R++}.
Nevertheless, GANs still suffer from poor generalizability and training stability, preventing them from scaling to unconstrained objects and scenes.
Recently, Diffusion Models (DMs)~\cite{FirstDiffusionModel,DDPM} have achieved great success in general 2D image synthesis, and are also applied to 3D~\cite{nichol2022pointe,jun2023shape,TriplaneDiffusion,HoloDiffusion,Rodin3DDM,HyperDiffusion,Train3DDM3,Train3DDM4}.
Yet, these methods train 3D DMs from scratch on specific objects such as human faces or vehicles, limiting their generalization.
Closer to ours are methods that adapt large-scale pre-trained 2D DMs~\cite{stable_diffusion} for 3D generation, which we will detail next.

\heading{Novel View Synthesis (NVS) with 2D Diffusion Models.}
Instead of reconstructing the entire 3D shape, NVS aims to generate 3D consistent images conditioned on a few input views~\cite{synsin,sajjadi2022scene}.
Early methods leverage the knowledge of epipolar geometry to perform interpolation between different input views~\cite{viewinterpolation,modelingandrendering,OldNVS1,OldNVS2}.
Since NVS is a 2D image-to-image translation task, recent works have re-purposed 2D DMs for it~\cite{gu2023nerfdiff,watson2022novel,TrainNVSDM1,GeNVS,TrainNVSDM2,TrainNVSDM3}.
To achieve 3D consistency, SparseFusion~\cite{zhou2023sparsefusion} builds a view-conditioned DM on the latent space of Stable Diffusion~\cite{stable_diffusion}, and utilizes Epipolar Feature Transformer (EFT)~\cite{EpipolarFeatureTransformer} to fuse features from input views.
Zero-1-to-3~\cite{zero1to3} directly fine-tunes Stable Diffusion on multi-view images rendered from Objaverse~\cite{deitke2022objaverse}.

The concurrent work MVDream~\cite{MVDream} proposes to denoise four views jointly with multi-view self-attention layers. However, camera pose information is fed in as 1D features to these models, discarding the explicit constraint of 3D geometry.
Thus this method does not allow accurate camera control.
To cope with this issue MVDream~\cite{MVDream} generates views at fixed camera positions spread apart 90 degrees from each other.
However, this approach limits the maximum number of views that can be generated to only 4.
Moreover, it limits the training data to only synthetic 3D model datasets, such as Objaverse~\cite{deitke2022objaverse}, since it requires rendering with the same fixed camera view for each object.

Other works thus study more explicit pose conditioning.
MVDiffusion~\cite{Mvdiffusion} derives inter-view dense correspondence from homography transformation, which is used to guide the attention module in Stable Diffusion.
iNVS~\cite{iNVS} applies image wrapping based on depth maps to re-use pixels from the source view, and thus only needs to inpaint occluded regions in novel view images.
While it can produce precise reconstructions when good depth maps are available, the quality of this method degrades drastically when depth maps are noisy or inaccurate. 
In addition, the depth ambiguity caused by epipolar lines used in iNVS results in the flipped prediction issue, where the model cannot differentiate two views from opposite directions.
SyncDreamer~\cite{SyncDreamer} instead builds a 3D feature volume by up-projecting features from each view, and then re-projects it to ensure 3D consistency among views.
However up-projection operation requires the network to explicitly understand the depth of each pixel, sharing the same issue with iNVS~\cite{iNVS}.

Different from prior works, we exploit the internal properties of large-scale pre-trained text-to-image diffusion models and enrich self-attention maps with the cross-view interactions derived from epipolar geometry.
In addition, we use Plücker coordinates~\cite{plucker} as positional encodings to inject 3D priors of the scene into the diffusion model, further improving camera conditioning and disambiguating different sides of the object.

\begin{figure*}[t]
    \vspace{\pagetopmargin}
    \vspace{-3mm}
    \centering
    \includegraphics[width=\linewidth]{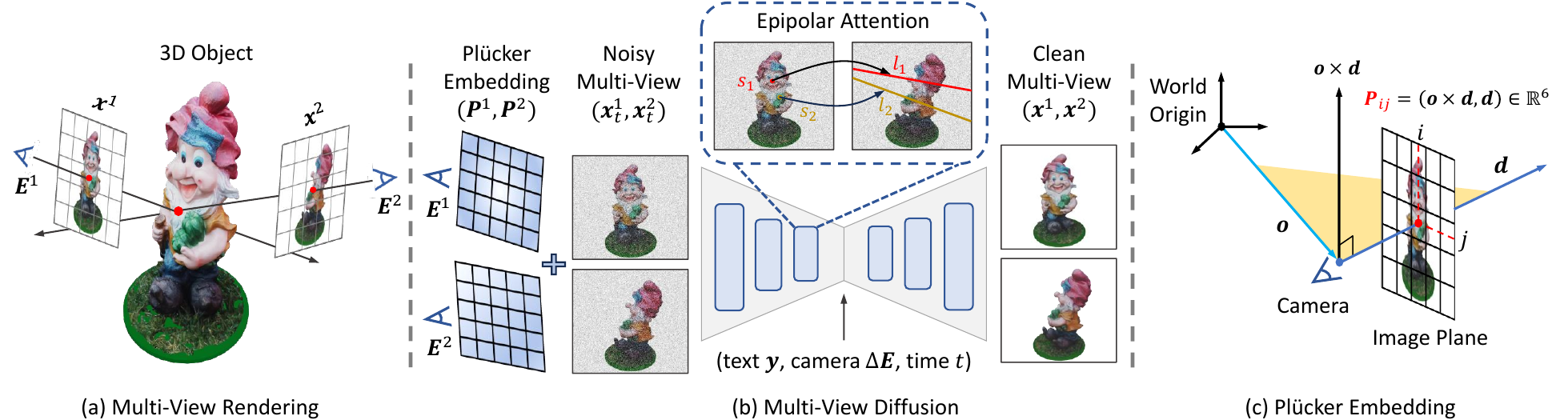}
    \vspace{-5mm}
    \caption{
        \textbf{Model pipeline.}
        (a) We initialize our multi-view diffusion model from pre-trained text-to-image model, and fine-tune it on multi-view renders of 3D objects.
        (b) Our model performs joint denoising on noisy multi-view images $\{\img^i_t\}_{i=1}^N$ conditioned on text $\textemb$ and relative camera poses $\deltacam$.
        Here, we illustrate the pipeline using $N=2$, which can be easily extended to more views.
        To enable cross-view interaction, we apply 3D self-attention by concatenating all views, and enforce epipolar constraints on the attention map.
        We further add (c) Plücker Embedding $\{\pluemb^i\}_{i=1}^N$ to the attention layers as positional encodings, to enable precise camera control and prevent object flipping artefacts (as shown in ~\cref{fig:ablation-qual}).
    }
    \vspace{\figmargin}
    \label{fig:model-pipeline}
\end{figure*}

\heading{Lifting 2D Diffusion Models to 3D Generation.}
Instead of training a model on 3D data, several works adopt a per-instance optimization paradigm where pre-trained 2D DMs provide image priors~\cite{dreamfusion,SJC}.
Some of them apply it for single-view 3D reconstruction~\cite{RealFusion,Magic123,dreambooth3d,shen2023anything3d,Make-It-3D,Nerdi,NeuralLift-360}.
More relevant to ours are text-to-3D methods that optimize a NeRF model~\cite{NeRF} by distilling the pre-trained text-to-image DM.
Follow-up works have improved this text-to-3D distillation process in many directions, including more efficient 3D representations~\cite{fantasia3d,TextMesh,magic3d,SDS-Explicit3DRep,Hd-fusion-SDS-DMTet}, better diffusion process~\cite{IT3D-SDS-OptwGAN,DreamTime-SDS-BetterOpt,DITTO-NeRF-SDS-ViewSampleOpt}, new loss functions~\cite{ProlificDreamer,HiFA-SDS-NewLoss}, and prompt design~\cite{SDS-NegPrompt}.
However, these methods still suffer from low visual quality and view consistency issues such as multi-face Janus and content drifting.
\algoNameFull generates multi-view images from a text prompt or a single input view with better 3D consistency and visual quality, which can mitigate these issues with multi-view distillation~\cite{MVDream}.

%% file: sections/3_method.tex
\section{Method}\label{sec:method}

\heading{Task Formulation.}
Our goal is to generate consistent and many novel views of the same object given a text prompt or an image, along with relative camera poses as input.
Towards this goal, we train a multi-view diffusion model that is made spatially aware by explicitly encoding the 3D knowledge of the scene.

We build upon a state-of-the-art 2D text-to-image diffusion model (\cref{sec:dm-review}).
Our specific adaptations enable 3D-aware interactions between views (\cref{sec:mv-diffusion}), which include 3D self-attention (\cref{sec:mv-diffusion-sa}), Epipolar Attention (\cref{sec:epipolar}), and Plücker Embeddings (\cref{sec:plucker}).

\subsection{Preliminary: Text-to-Image Diffusion Models}\label{sec:dm-review}

Diffusion models (DMs)~\citep{DDPM,FirstDiffusionModel} are generative models that learn a target data distribution $p_\theta(\img_0)$ by gradually denoising a standard Gaussian distribution, denoted as $p_\theta(\img_0) = \int p_\theta(\img_{0:T})\ d\img_{1:T}$, where $\img_{1:T}$ are intermediate noisy samples.
DMs leverage a forward process that iteratively adds Gaussian noise $\noise$ to the clean data $\img_0$, which is controlled by a pre-defined variance schedule $\{\bar{\alpha}_t\}_{t=1}^\top$.
During training, we manually construct noisy samples $\img_{t} = \sqrt{\bar{\alpha}_t} \img_0 + \sqrt{1 - \bar{\alpha}_t} \noise_t$, and train a denoiser model $\dmmodel(\img_{t}, t)$ to predict the added noise conditioned on the denoising time step $t$:
\begin{equation}\label{eq:dm-loss}
    \mathcal{L}_{\mathrm{DM}} = ||\noise_t - \dmmodel(\img_{t}, t)||^2,\ \mathrm{where\ \ } \noise_t \sim \mathcal{N}(\bm{0}, \bm{I}).
\end{equation}
Generally, the denoiser $\dmmodel$ is parameterized as a U-Net~\cite{unet}, which comprises of interleaved residual blocks~\cite{ResNet} and self-attention layers~\cite{Attention}.
Within this U-Net, we are interested primarily in self-attention layers~\cite{Attention}, and we refer the reader to the original paper~\cite{stable_diffusion} for the overview of other blocks. 
The self-attention layer takes as input a feature map $\featmap$ and compute attention of feature in location s with entire feature map:
\begin{equation}\label{eq:cross-attn}
    \tilde{\featmap}_\srcpoint = \text{SoftMax}\left(\frac{Q(\featmap_\srcpoint) \cdot K(\featmap)^\top}{\sqrt{d}}\right) \cdot V(\featmap),
\end{equation}
where $Q, K, V$ are linear projection layers, $\featmap \in \mathbb{R}^{(hw) \times d}$ is a flattened feature map obtained from the 2D denoiser $\dmmodel$, where $d$ is the feature dimension, and $h,w$ are intermediate spatial dimensions. $\featmap_\srcpoint$, $\tilde{\featmap}_\srcpoint$ is the input and output feature for location $\srcpoint$ respectively.
In practice, the self-attention operation occurs at multiple resolutions in $\dmmodel$.

\subsection{Multi-View Diffusion Models}\label{sec:mv-diffusion}

Inspired by the success of text-to-image DMs, we propose to generate multi-view images by fine-tuning a pre-trained 2D DM on multi-view rendered images of 3D assets.
\cref{fig:model-pipeline} shows the overall pipeline of our framework, \algoNameFull.
In this section, we use $N=2$ views to explain our method for brevity.
However, note that \algoNameFull is easily extensible to generate an arbitrary number of views.

\subsubsection{Multi-View Self-attention}\label{sec:mv-diffusion-sa}

The goal of our multi-view DM $\dmmodel(\img_t^1, \img_t^2, t, \textemb, \deltacam)$ is to generate 3D consistent images $(\img^1, \img^2) \in \mathbb{R}^{H \times W \times 3}$  of an object guided by a text input $\textemb$ and their relative camera pose $\deltacam  \in \mathbb{R}^{3 \times 4}$.
To enable cross-view interaction, we concatenate the feature maps of two views side-by-side as input to the self-attention layers, denoted as $[\featmap^1|\featmap^2]$.
This allows each location $\srcpoint$ on $\featmap^1$ to attend to all locations on itself and $\featmap^2$, calculated as:
\begin{equation}\label{eq:vanilla-3d-attn}
    \tilde{\featmap}_\srcpoint^1 = \text{SoftMax}\left(\frac{Q(\featmap_\srcpoint^1) \cdot K([\featmap^1|\featmap^2])^\top}{\sqrt{d}}\right) \cdot V([\featmap^1|\featmap^2]).
\end{equation}

\begin{figure}[t]
    \vspace{\pagetopmargin}
    \vspace{-2mm}
    \centering
    \includegraphics[width=\linewidth]{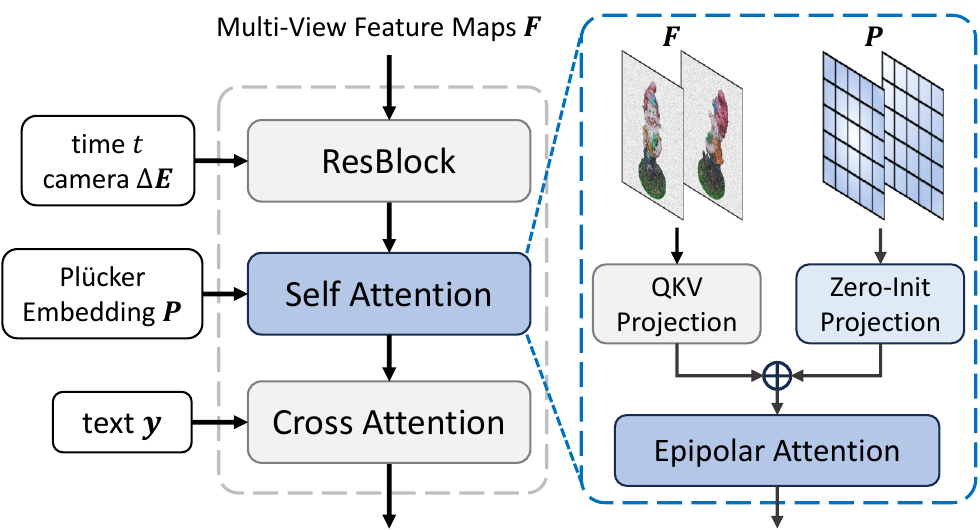}
    \vspace{\figcapmargin}
    \vspace{-3mm}
    \caption{
        \textbf{Illustration of one block in our multi-view diffusion model}, which consists of a residual block, a self-attention layer, and a cross-attention layer.
        The residual block guides the model on the denoising timestep $t$ and the relative camera pose $\deltacam$, while the cross-attention layer conditions on text $\textemb$.
        We add Plücker Embedding $\pluemb$ to feature maps $\featmap$ in the self-attention layer by inflating the original QKV projection layers with zero projections.
    }
    \vspace{\figmargin}
    \label{fig:unet-block}
\end{figure}

\heading{Camera conditioning.}
We embed the relative camera pose $\deltacam$ with an MLP and fuse it with timestep embedding of DM to condition the residual blocks as shown in \cref{fig:unet-block}, similar to \cite{MVDream}.

\heading{Issues with vanilla self-attention.}
We find empirically that such an unconstrained multi-view self-attention leads to content-copying between views (shown in Figure~\ref{fig:ablation-qual}), \ie the model generates similar images when the camera pose difference $\deltacam$ is small, ignoring the underlying 3D geometry.
We hypothesize that this could be the reason concurrent works such as MVDream~\cite{MVDream} opt to generate images with 90 degree view change (along azimuths and fixed elevation) -- as it diminishes the overlap between different views.

\begin{figure}[t]
    \vspace{\pagetopmargin}
    \vspace{-3mm}
    \centering
    \includegraphics[width=0.9\linewidth]{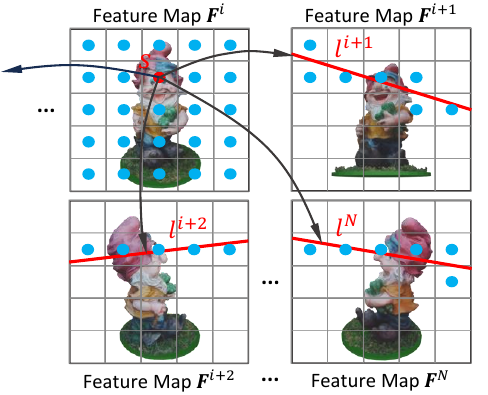}
    \vspace{\figcapmargin}
    \caption{
        \textbf{Epipolar Attention.}
        For each point $\srcpoint$ ({\color{red} red} point) on a feature map $\featmap^i$, we compute its epipolar lines $\{\epiline^{j}\}_{j \ne i}$ on all other views $\{\featmap^j\}_{j \ne i}$.
        Point $\srcpoint$ will only attend to features along these lines plus all the points on itself ({\color{cyan} blue} points).
    }
    \vspace{\figmargin}
    \label{fig:epi-attn}
\end{figure}

\subsubsection{Multi-View Epipolar Attention}\label{sec:epipolar}

To enable \algoNameFull to synthesize views at arbitrary relative poses, and address the above content-copying challenge, we propose to replace the vanilla self-attention operation with Epipolar Attention~\cite{EpipolarFeatureTransformer}.
Epipolar Attention works by restricting the positions a point in feature map can attend to in other views -- by exploiting epipolar geometry.
\cref{fig:epi-attn} presents this mechanism.
Specifically, given a source point $\srcpoint$ on a feature map $\featmap^i$, we compute its epipolar lines (implemented as a set of points) $\{\epiline^j\}_{j \ne i}$ on all the other views $\{\featmap^j\}_{j \ne i}$.
When computing the attention map between views, we ignore points that do not lie on these epipolar lines, so that the source point $\srcpoint$ only has access to features that lie along the camera ray (in other views) as well as all points in its own view for denoising:
\begin{equation}\label{eq:epipolar-3d-attn}
    \tilde{\featmap}_\srcpoint^i = \text{SoftMax}\left(\frac{Q(\featmap_\srcpoint^i) \cdot K([\featmap^i|\featmap^j_{\epiline^j}])^\top}{\sqrt{d}}\right) \cdot V([\featmap^i|\featmap^j_{\epiline^j}]).
\end{equation}
In practice, we dilate the epipolar lines with a 3$\times$3 filter to consider neighboring target points for better robustness.
Overall, Epipolar Attention enhances our generalizability to unseen viewpoint differences and objects.

\heading{Issues with Epipolar Attention.}

However, solely using epipolar lines to constrain attention masks models can cause flipped 
predictions especially under large viewpoint changes.
This happens because in the absence of precise depth of the object surface, the model can leverage information from any point along these lines.
Consider the feature map $\featmap^N$ in \cref{fig:epi-attn}, the source point $\srcpoint$ may attend to the front face of the figure or the back of it.
The latter will cause a flipped prediction.
iNVS~\cite{iNVS} solves this problem with a monocular depth estimator.
Yet, their imprecise depth leads to deformed object surfaces and distorted textures.
We instead address this issue with a Plücker Ray Embedding which is detailed next.

\subsubsection{Plücker Ray Embedding}\label{sec:plucker}

Given a camera with its center placed at $\camorigin \in \mathbb{R}^3$ in the world coordinate system, we represent a ray passing through a point on the feature map $\featmap_{ij}$ along a normalized direction $\camdir \in \mathbb{R}^3$ as $\ray_{ij}$.
We embed this ray as the positional encoding to help our model distinguish between different views. 
We find the simple ray parametrization $\ray_{ij} = (\camorigin, \camdir)$ to be insufficient here. 
As an example, consider two rays in the same direction but with different camera origins which lie along this ray, $\ray_{ij}^1 = (\camorigin, \camdir)$ and $\ray_{ij}^2 = (\camorigin + t \camdir, \camdir)$, their embeddings turn out to be notably different, despite them representing essentially the same ray.

Inspired by recent works in Neural Light Fields~\cite{LFNs,RayConditioningGAN}, we adopt the Plücker Ray Embedding $\pluemb_{ij} = (\camorigin \times \camdir, \camdir)$ where $\times$ is the cross product.
See \cref{fig:model-pipeline} (c) for illustration.
This parametrization is able to map $\ray^1$ and $\ray^2$ to the same embedding as we have:
\begin{equation}\label{eq:plucker-eq}
    (\camorigin + t \camdir) \times \camdir = \camorigin \times \camdir + t \camdir \times \camdir = \camorigin \times \camdir.
\end{equation}
We simply pass the Plücker Embedding $\pluemb \in \mathbb{R}^{(hw) \times 6}$ through a linear projection layer to project it into $d$ dimension and add it to the multi-view feature maps $\featmap$, which serves as the input to the Epipolar Attention layer, shown in \cref{fig:unet-block}.
To avoid disturbing the pre-trained model, the weights of the projection layer are initialized as zeros and learned during fine-tuning similar to ControlNet~\cite{zhang2023adding}.

\heading{3D geometric priors in Plücker Embedding.}
With the Plücker coordinates, rays that are close in the 3D space share similar embeddings, which leads to higher values in the pre-softmax self-attention map $Q(\featmap) \cdot K(\featmap)^\top$.
This encourages feature points to look at spatially nearby locations in other views, enhancing the 3D consistency across views. %
On the other hand, two rays passing through the same 3D location from opposite cameras will have Plücker coordinates with flipped (positive/negative) sign. Their embeddings will have a smaller dot product and result in a smaller attention value.
The two pixels will thus attend less to each other, effectively addressing the flipped prediction problem.

%% file: sections/4_experiment.tex
\section{Experiments}

We conduct extensive experiments to answer the following questions:
\textbf{(i)} Can \algoNameFull generate high-quality multi-view images from diverse (non-orthogonal and overlapping) viewpoints that are aligned with input text or image? (\cref{sec:text-to-mv})
\textbf{(ii)} Are the synthesized views 3D consistent? (\cref{sec:img-to-mv})
\textbf{(iii)} To what extent do Plücker Positional Embeddings and Epipolar Attention contribute to the overall performance? (\cref{sec:img-to-mv})
\textbf{(iv)} Lastly, can \algoNameFull enable high-quality text-guided 3D asset generation? (\cref{sec:text-to-3d-sds})

\subsection{Experimental Setup}\label{sec:exp-setup}

\heading{Training Data Curation.}
Instead of using the entire Objaverse~\cite{deitke2022objaverse} which consists of many flat and primitive shapes that can drift the diffusion model away from high-quality generation, we filter Objaverse using a few simple heuristics based on its metadata. We use captions from Cap3D~\cite{luo2023scalable}. 
We select 150K objects with the most like, view, and comment count available in metadata, as well as the top 50K objects that contain the highest number of mesh polygons and vertex counts.
We use Blender~\cite{Blender} to render 12 multi-view images for each object at a resolution of 256$\times$256.
All objects are centered and re-scaled to a unit cube.
We randomly sample camera positions with elevations in $[-90\degree, 90\degree]$, azimuths in $[0\degree, 360\degree]$, and fix the distance to origin as 3.5 and FOV as 40.26$\degree$.

\heading{Training Details.}
We initialize \algoNameFull from the pre-trained weights of Stable Diffusion v1.5~\cite{stable_diffusion}.
We train two versions of our model: one with text conditioning and another one with image conditioning for the novel view synthesis task.
In both the variants, we set the number of views $N$ to 2 while training.
For the text-conditioned model, we jointly denoise both the views.
For the image-conditioned model, we feed in one clean source view image and denoise the target view.
All our baselines and reported numbers follow this setup. First, we train our models two-view models for 40K iterations on Objaverse, with an effective batch size of 1728 samples per iteration, on eight H100 GPUs. 
Later we train a larger text-conditioned model with $N=4$ views on sixteen H100 GPUs for 100K steps, and use it to generate all visuals (except ablation study). 

\heading{Evaluation Datasets and Metrics.}
For text-conditioned multi-view image generation, we follow MVDream~\cite{MVDream} and randomly sample 1,000 Objaverse captions as text prompts to generate images.
We use CLIP-score~\cite{radford2021learning} to measure the image-text alignment.
We also report the Inception Score (IS)~\cite{InceptionScore} and Fréchet Inception Distance (FID)~\cite{FID} of the generated images to evaluate image generation quality. It is important to highlight that these metrics only measure generation quality of individual images, and do not provide any information about their multi-view 3D consistency. 

For image-conditioned novel view synthesis, we select 1,000 unseen Objaverse objects which are not contained in our training set for testing.
Following \cite{zero1to3,iNVS}, we also adopt real-world scanned objects from the \textit{Google Scanned Objects} (GSO) dataset~\cite{downs2022google} to evaluate the generality of our method.
We render each object from two views following the same setup in training data generation, where one view serves as the model input and another view is the target novel view image. We report PSNR, SSIM~\cite{ssim}, and LPIPS~\cite{LPIPS} metrics to measure the accuracy of the synthesized novel views.

\heading{Baselines.}
Since it is difficult to replicate and control for training and rendering setups used in prior works, we choose the following variants of our model as primary baselines:
\textit{Vanilla MV-DM} that only adds 3D self-attention on concatenated multi-view feature maps without Epipolar Attention and Plücker Embedding;
\textit{MV-DM (Epipolar)} and \textit{MV-DM (Plücker)} which incorporate the two components, respectively.
We also compare \algoNameFull with two \textit{concurrent} works: \textit{MVDream}~\cite{MVDream} and \textit{SyncDreamer}~\cite{SyncDreamer}.
Different from \algoNameFull, both methods can only generate views at a fixed elevation and azimuth ranges.
In the image-conditioned novel view synthesis task, we compare with additional baselines \textit{Zero-1-to-3}~\cite{zero1to3} and \textit{iNVS}~\cite{iNVS}.
We used the official codebase and pre-trained weights of these methods on our testing data to report their results.

\begin{table}[t]
    \centering
    \vspace{\pagetopmargin}
    \setlength{\tabcolsep}{6pt}
    \small
    \begin{tabular}{lHcc}
        \toprule
        \textbf{Method} & FID $\downarrow$ & IS $\uparrow$ & CLIP-score $\uparrow$ \\
        \midrule
        \color{gray} MVDream (v2.1) $^\dagger$~\cite{MVDream} & \color{gray} 32.06 & \color{gray} \pms{13.36}{0.87} & \color{gray} \pms{30.22}{3.83} \\
        \midrule
        MVDream (v1.5) $^\dagger$~\cite{MVDream} & - & \pms{9.72}{0.43} & \pms{28.55}{4.05} \\
        SyncDreamer $^\ddagger$~\cite{SyncDreamer} & - & \pms{11.69}{0.24} & \pms{27.76}{4.84} \\
        \midrule
        Vanilla MV-DM & 55.25$^{*}$ & \pms{11.04}{0.81} & \pms{28.52}{3.69} \\
        \algoNameFull (\textbf{Ours}) & xxx & \pms{11.18}{0.97} & \pms{29.87}{3.33} \\
        \bottomrule
    \end{tabular}
    \vspace{\tablecapmargin}
    \caption{
        \textbf{Quantitative results on text-conditioned multi-view image generation.}
        We randomly sample 1,000 captions from Objaverse, and evaluate the FID, Inception Score (IS), and CLIP-score.
        $^\dagger$ We ran MVDream's code on the same captions we used.
        $^\ddagger$ We first generated single-view images using Stable Diffusion~\cite{stable_diffusion} on the same captions we used and removed their backgrounds.
        Then, we ran SyncDreamer's code to generate multi-view images.
    }
    \label{tab:text-to-mv-2d-metrics}
    \vspace{\tablemargin}
\end{table}

\subsection{Text-conditioned Multi-View Generation}\label{sec:text-to-mv}

We use single view quality metrics to compare methods, similar to MVDream. 
We evaluate two MVDream variants, which are fine-tuned from Stable Diffusion v1.5 (same as ours) and v2.1, respectively.
For SyncDreamer, we follow the text-to-image-to-3D pipeline described in their paper to first generate a single-view image from a text prompt using Stable Diffusion, and then generate multiple views from it.
We make sure that the single-view image is aligned with the text, and pre-process it using the script provided in their official codebase. 

\heading{\algoNameFull is a strong 2D text-to-image generator.}
The results on image generation quality are presented in \cref{tab:text-to-mv-2d-metrics}. 
\algoNameFull outperforms or matches both baselines on 2D Image Quality metrics when compared against the methods utilizing the same underlying Stable Diffusion v1.5 base model.
This confirms that our method while being more 3D consistent, does not compromise either text-to-image alignment or overall image quality, but rather improves it compared to our baseline MV-DM. 

We provide qualitative results in \cref{fig:teaser}, \cref{app-fig:supp-ours-cfg}, and \cref{app-fig:supp-ours-cfg-sd21}. \algoNameFull is able to generate consistent multi-view images of diverse 3D subjects, ranging from daily objects to highly complex machines. {Additionally, we put preliminary investigations of training SPAD with v2.1 base model in \cref{app-sec:v21-weights}.}

\begin{table}[t]
    \centering
    \vspace{\pagetopmargin}
    \setlength{\tabcolsep}{5pt}
    \small
    \begin{tabular}{lccc}
        \toprule
        \textbf{Method} & PSNR $\uparrow$ & SSIM $\uparrow$ & LPIPS $\downarrow$ \\
        \midrule
        Zero-1-to-3 $^\dagger$~\cite{zero1to3} & 18.16 & 0.81 & 0.201 \\
        iNVS~\cite{iNVS} & \textbf{20.52} & 0.81 & 0.178 \\
        SyncDreamer $^\dagger$~\cite{SyncDreamer} & 19.51 & \textbf{0.84} & \underline{0.174} \\
        \midrule
        Vanilla MV-DM & 17.56 & 0.81 & 0.20 \\
        MV-DM (Epipolar) & 18.90 & \underline{0.82} & 0.19 \\
        MV-DM (Plücker) & 17.98 & 0.81 & 0.20 \\
        \algoNameFull (\textbf{Ours}) & \underline{20.29} & \textbf{0.84} & \textbf{0.166} \\
        \bottomrule
    \end{tabular}
    \vspace{\tablecapmargin}
    \caption{
        \textbf{Quantitative results on image-conditioned novel view synthesis on Objaverse.}
        We report PSNR, SSIM, and LPIPS on the generated novel view images of 1,000 unseen Objaverse objects.
        $^\dagger$ We run the official codebases of SyncDreamer and Zero-1-to-3 to report results.
    }
    \label{tab:nvs-3d-metrics-objaverse}
    \vspace{\tablemargin}
\end{table}

\begin{table}[t]
    \centering
    \setlength{\tabcolsep}{5pt}
    \small
    \begin{tabular}{lccc}
        \toprule
        \textbf{Method} & PSNR $\uparrow$ & SSIM $\uparrow$ & LPIPS $\downarrow$ \\
        \midrule
        Zero-1-to-3~\cite{zero1to3} & 16.10 & 0.82 & 0.183 \\
        iNVS~\cite{iNVS} & \textbf{18.53} & 0.80 & 0.180 \\
        SyncDreamer $^\dagger$~\cite{SyncDreamer} & 17.18 & \textbf{0.83} & \underline{0.178} \\
        \midrule
        Vanilla MV-DM & 15.98 & {0.81} & 0.20 \\
        MV-DM (Epipolar) & 17.13 & \underline{0.82} & 0.19 \\
        MV-DM (Plücker) & 16.15 & {0.81} & 0.20 \\
        \algoNameFull (\textbf{Ours}) & \underline{17.99} & \textbf{0.83} & \textbf{0.169} \\
        \bottomrule
    \end{tabular}
    \vspace{\tablecapmargin}
    \caption{
        \textbf{Quantitative results on image-conditioned novel view synthesis on GSO.}
        We report PSNR, SSIM, and LPIPS on the generated novel view images of GSO objects.
        $^\dagger$ SyncDreamer only reports results on 30 selected objects from GSO in their paper~\cite{SyncDreamer}.
        We ran their code and test it on all the GSO objects here.
    }
    \label{tab:nvs-3d-metrics-gso}
    \vspace{\tablemargin}
\end{table}

\subsection{Image-conditioned Novel View Synthesis}\label{sec:img-to-mv}

Since image quality metrics do not provide any indication of multi-view consistency or the quality of camera control. For evaluation of multi-view consistency, we mostly rely on image-conditioned experiments.
For this evaluation, given an input view and relative camera pose, we generate the target view and compare it against ground truth. 
\cref{tab:nvs-3d-metrics-objaverse} and \cref{tab:nvs-3d-metrics-gso} present the novel view synthesis results on Objaverse and GSO, respectively. 

\heading{\algoNameFull preserves structural and perceptual details faithfully.} We find that \algoNameFull outperforms all baselines on LPIPS metrics across both datasets, while matching its performance to SyncDreamer on SSIM. Moreover, we find that adding each component (Epipolar and Plücker) gradually improves scores across the board, and leads to state-of-the-art performance with our full model. This confirms our main hypothesis that imparting 3D understading to MV-DMs is helpful. 

We also find that iNVS~\cite{iNVS} is able to achieve the highest PSNR, since it directly copies pixels from the source view (via depth-based reprojection).
However, it particularly performs worse on SSIM and LPIPS metrics, which measure the structural and semantic accuracy of the generated view.
This is because of deformations introduced by reprojection when viewpoint changes are large and monocular depth from ZoeDepth~\cite{bhat2023zoedepth} is inaccurate. 

The official inference code of SyncDreamer always generates 16 views at fixed azimuth angles uniformly distributed in $[0\degree, 360\degree]$, which is incompatible with our random view generation setup.
We modified their code to consider the exact target camera pose as model input, but found it performed worse than choosing the prediction at the azimuth that is closest to the target azimuth.
Therefore, we report SyncDreamer results using the closest view, where the error is usually smaller than $10\degree$.

\begin{figure*}[t]
    \vspace{\pagetopmargin}
    \centering
    \includegraphics[width=\linewidth]{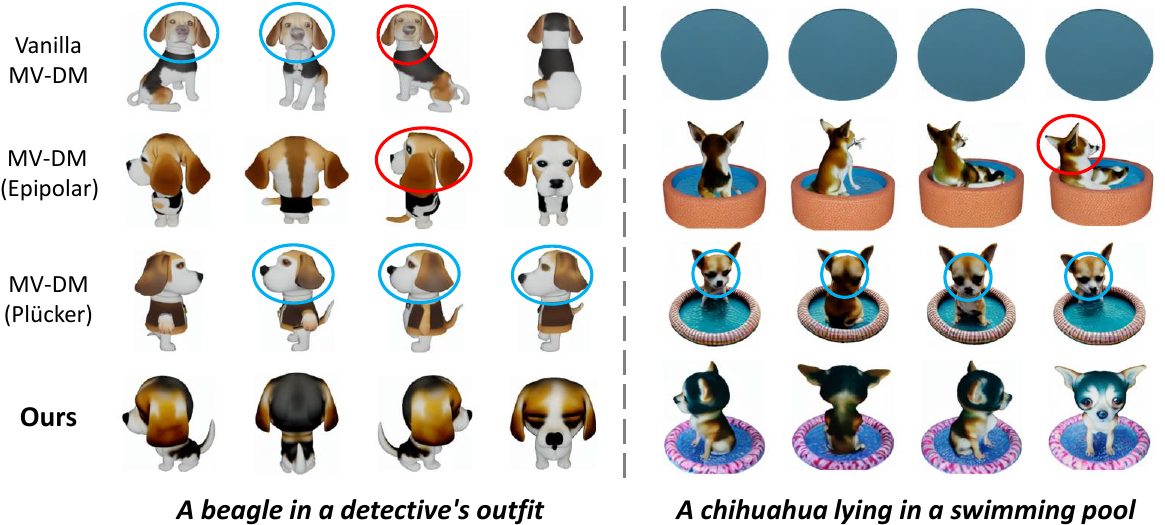}
    \vspace{-5mm}
    \caption{
        \textbf{Qualitative comparison between \algoNameFull and its variants.}
        We prompt models trained on two views to generate four views at 90 degree intervals for clear visual distinctions.
        The flipped predicted views are highlighted with {\color{red} red} circles, while the content-copying issues are indicated by {\color{cyan} blue} circles.
    }
    \vspace{\figmargin}
    \label{fig:ablation-qual}
\end{figure*}

\subsection{Qualitative Analysis}
We also conduct qualitative analysis to visually understand the usefulness of each component of our model. The text-conditioned multi-view generation results of baselines and \algoNameFull are shown in \cref{fig:ablation-qual}, where all models are trained with two views while prompted to generate four views.
The elevation is fixed for all the views, and the azimuth spans uniformly between $[0\degree, 360\degree]$.

\heading{Epipolar Attention promotes better camera control in \algoNameFull.}
We find that the vanilla (full) 3D self-attention used in Vanilla MV-DM and MV-DM (Plücker) models often leads to content copying.
This is highlighted in the figure using blue circles, where the generated dogs face in similar direction, ignoring the target camera poses. We hypothesize that the readaptation of the self-attention layer of SD originally trained to attend only to itself hinders with generalization and controllability of this model. 

Additionally, since these models are trained only to generate two views, we hypothesize that they overfit to predicting only two novel views. In contrast, Epipolar Attention constrains cross-view interactions to only happen between spatially related pixels, reducing the search space in establishing correspondences across images. Despite not being trained on four views, the model is still able to generate 3D consistent images by attending to the correct regions.

\heading{Plücker Embeddings help prevent generation of flipped views.}
When the difference in camera positions between two views is large, the epipolar lines introduce ambiguities in the ray directions.
Indeed, Vanilla MV-DM and MV-DM (Epipolar) sometimes predict image regions that are rotated by $180\degree$.
For example, the dog's head highlighted by red circles looks in the opposite direction of the body, which is inconsistent with other views.
Instead, Plücker Embeddings bias the model to pay less attention to camera views on opposite sides of the object, while leveraging more information from spatially closer views.

\subsection{Text-to-3D Generation}\label{sec:text-to-3d-sds}

\begin{figure*}[t]
    \vspace{\pagetopmargin}
    \centering
    \includegraphics[width=1.0\linewidth]{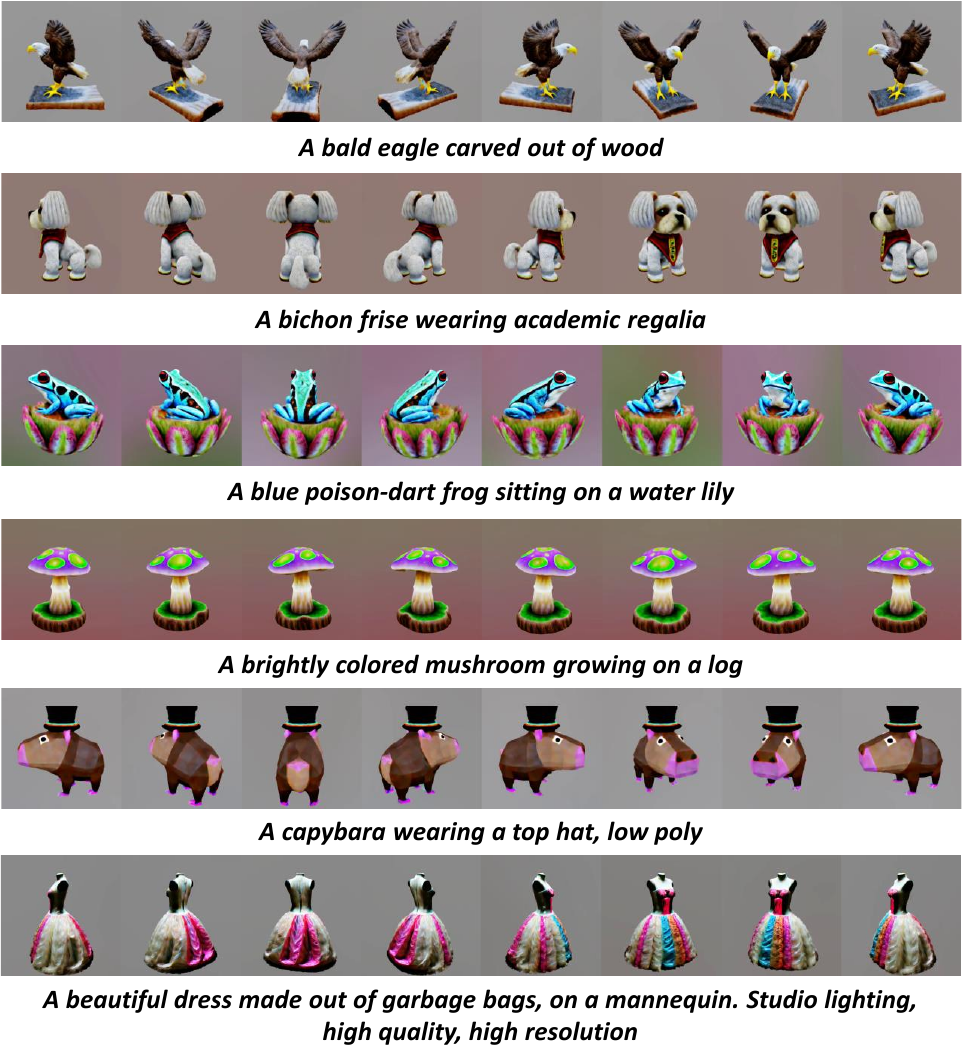}
    \vspace{\figcapmargin}
    \vspace{-3mm}
    \caption{
        \textbf{Text-to-3D generation using multi-view SDS with \algoNameFull.}
        We adopt the multi-view SDS proposed in MVDream~\cite{MVDream} to train a NeRF model.
        Thanks to the 3D consistency of our model, we do not suffer from the multi-face Janus issue.
    }
    \vspace{\figmargin}
    \label{app-fig:sds-3d-content-gen}
\end{figure*}

\heading{Multi-view SDS.} Inspired by \cite{zero1to3,MVDream}, we also adopt the multi-view Score Distillation Sampling (SDS)~\cite{dreamfusion} to perform text-to-3D generation using the four-view \algoNameFull variant.
Concretely, we integrate our model into the state-of-the-art text-to-3D generation codebase threestudio~\cite{threestudio2023}, and follow the setup similar to MVDream~\cite{MVDream} for stable NeRF~\cite{NeRF} distillation. \cref{app-fig:sds-3d-content-gen} shows the multi-view rendered images of the trained NeRF models. We find that \algoNameFull is able to reconstruct consistent geometry without Janus problem, while maintaining good visual quality.

\begin{figure*}[t]
    \vspace{\pagetopmargin}
    \centering
    \includegraphics[width=1.0\linewidth]{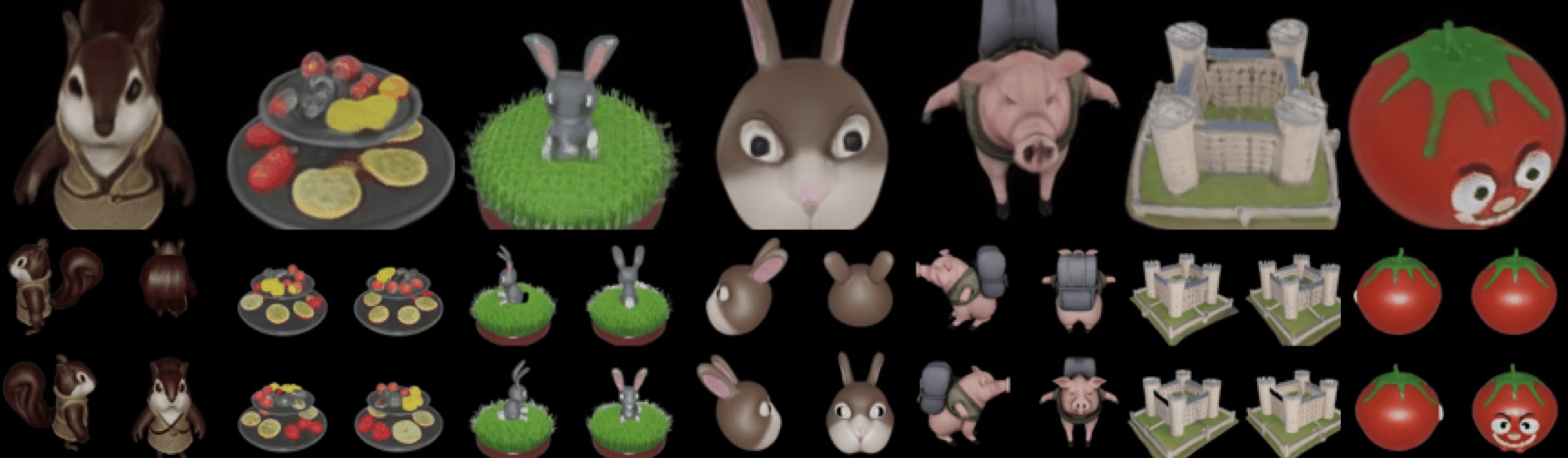}
    
    \includegraphics[width=1.0\linewidth]{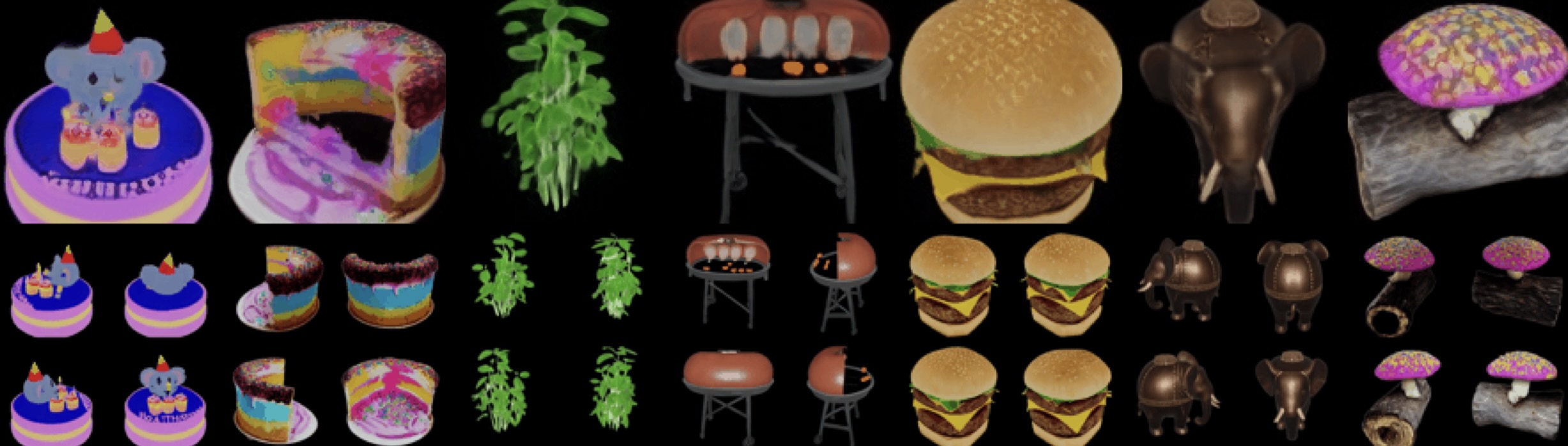}
    \caption{
        \textbf{Text-to-3D generation using multi-view Triplane generator with \algoNameFull.}
        Following~\cite{hong2023lrm, instant3d2023} we trained a multi-view conditioned triplane generator that outputs a NeRF using four outputs of \algoNameFull in a single feed-forward pass. We show the rendered NeRF on the top row (zoomed) and corresponding multi-view outputs from \algoNameFull in the bottom row. For entire 360-degree videos see our website. 
    }
    \vspace{\figmargin}
    \label{app-fig:triplane-3d-gen}
\end{figure*}

\heading{Multi-view Triplane Generator.} Inspired by concurrent works \cite{hong2023lrm, instant3d2023}, we trained a multi-view images to triplane generator on Objaverse. We follow closely followed the setup from Instant3D~\cite{instant3d2023}, and used four orthogonal views from \algoNameFull to generate a NeRF in a single feed-forward pass. Combined together this approach takes roughly 10 seconds to generate a single asset from text prompt, which is greater than two orders of magnitude faster than SDS optimization. \cref{app-fig:triplane-3d-gen} shows the results from this experiment. Thus, we find that \algoNameFull can be used as a faithful base model to facilitate such generations. 

%% file: sections/5_conclusion.tex
\section{Conclusion}
In this paper, we propose \algoNameFull, a novel framework for generating multiple views from text or image input.
We propose to transform the self-attention layers of the pre-trained text-to-image diffusion model into Epipolar Attention to promote multi-view interactions and improve camera control.
Moreover, we augment self-attention layers with Plücker positional encodings to further improve camera control by preventing flipping view prediction of the object.
We provide rigorous evaluations of these modifications and demonstrate state-of-the-art results in terms of image-conditioned novel view synthesis.

\heading{Limitations and Future Work}. While our method improves the 3D consistency of multi-view diffusion models, there still remains lots of scope for improvements. For example, a larger Stable Diffusion such as SDXL~\cite{podell2023sdxl} can further improve performance while preventing lossy compression of image conditioning. We can use a monocular depth estimator similar to iNVS~\cite{iNVS} to further improve the correspondences established by epipolar self-attention. Finally, we plan to explore the usage of \algoNameFull to generate dynamic 4D assets and multi-object scenes. Additionally, datasets of monocular videos and pre-trained text-to-video generators can be explored to improve the quality and consistency of the generated results.

%% file: sections/6_appendix.tex
\appendix

\section{More Implementation Details}

\heading{DDIM Initialization.} We use a black Gaussian-blob on a white background to initialize the first 20 steps (out of 200) of our DDIM sampler, which ensures that our model correctly generates a single object placed at the center on the white background (similar to images during training). This trick is similar to the one used in iNVS~\cite{iNVS} which starts inpainting with the partial image warped with depth.

We list our hyper-parameter choices and miscellaneous training details below: 

\begin{table}[h!]
\center
\begin{tabular}{Hlc}
\toprule
    \# & Hyper-parameters & Value \\
    \midrule
    {1} & Base Learning rate & 1e-4 \\
    {2} & Learning rate decay & None \\
    {3} & Loss Type & L2 \\ 
    {5} & Classifier-free guidance & 7.5 (text-only) \\
    {7} & Effective batch size & 1152 \\
    {8} & DDIM Steps & 200 \\
    {9} & Gaussian Blob Initialization Steps & 20 \\
    {11} & CLIP Frozen & True \\
    {12} & Renders background color & White \\
    {13} & Image Resolution & 256 \\
    {14} & Learning rate linear warmup & 100 steps \\ 
    \midrule
\end{tabular}
\vspace{\tablecapmargin}
\caption{Hyperparameter choices for \algoNameFull.} 
\label{tab:hyperparameters_supp}
\vspace{\tablemargin}
\end{table}

\section{Additional Experiments and Results}

\subsection{Qualitative Results and Baseline Comparisons}\label{app-sec:more-spad-results}

\begin{figure*}[t]
    \vspace{\pagetopmargin}
    \centering
    \includegraphics[width=1.0\linewidth]{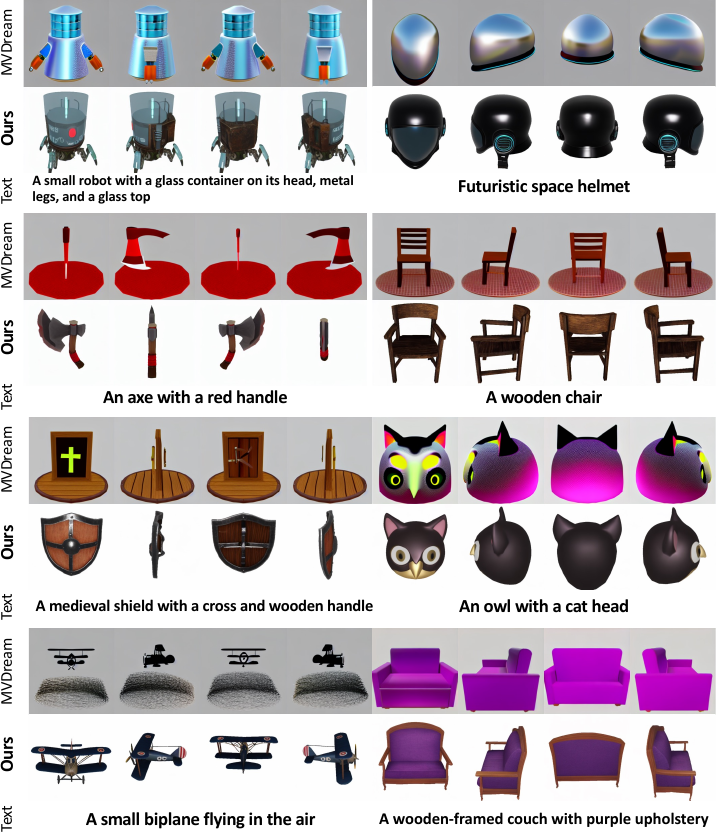}
    \vspace{\figcapmargin}
    \vspace{-3mm}
    \caption{
        \textbf{Comparison of text-conditioned multi-view generation with MVDream~\cite{MVDream}}.
    }
    \vspace{\figmargin}
    \label{app-fig:t2mv-vs-mvdream}
\end{figure*}

\heading{Text-conditioned multi-view generations and comparison with MVDream~\cite{MVDream}.}
\cref{app-fig:t2mv-vs-mvdream} presents the results.
We find \algoNameFull synthesizes images with higher quality details and better alignment with the text prompt.

\begin{figure*}[t]
    \vspace{\pagetopmargin}
    \centering
    \includegraphics[width=1.0\linewidth]{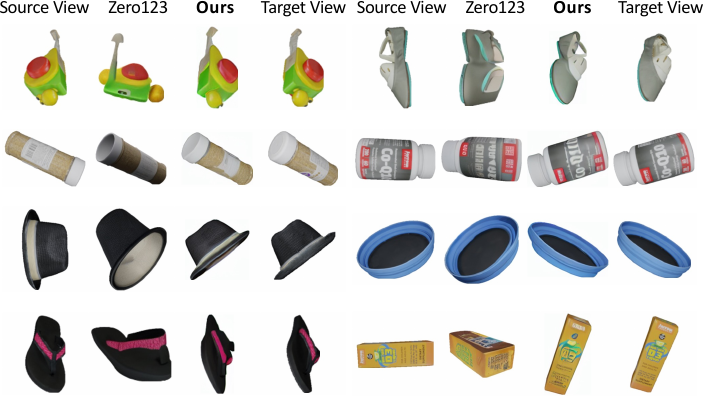}
    \vspace{\figcapmargin}
    \vspace{-3mm}
    \caption{
        \textbf{Comparison of image-conditioned novel view synthesis with Zero123~\cite{zero1to3}}.
    }
    \vspace{\figmargin}
    \label{app-fig:nvs-vs-z123}
\end{figure*}

\heading{Image-conditioned novel view synthesis and comparison with Zero123~\cite{zero1to3}.}
\cref{app-fig:nvs-vs-z123} presents the results.
We find \algoNameFull preserves the structural and perceptual details of objects and exhibits better 3D consistency.

\begin{figure*}[t]
    \vspace{\pagetopmargin}
    \centering
    \includegraphics[width=1.0\linewidth]{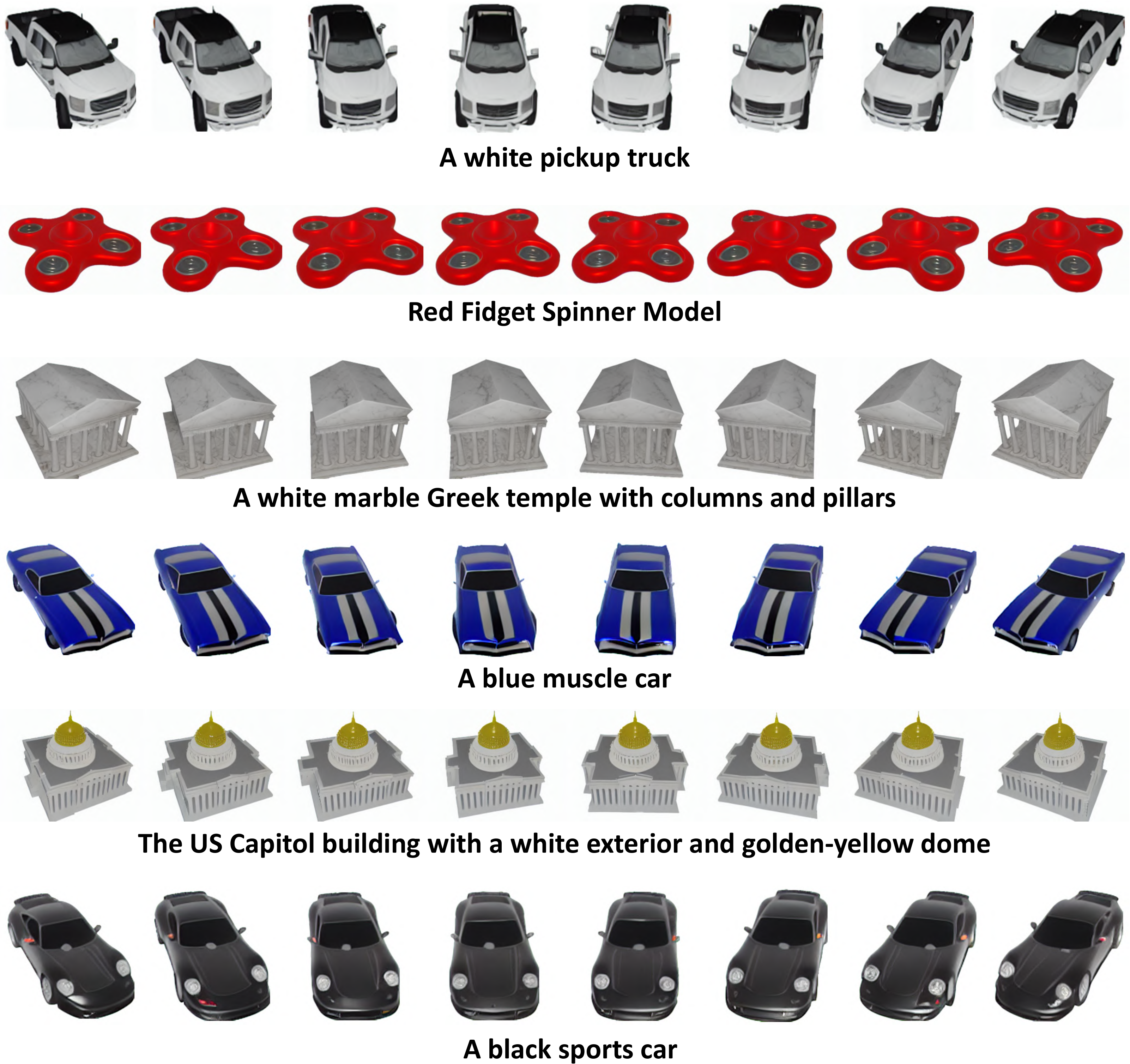}
    \vspace{\figcapmargin}
    \vspace{-3mm}
    \caption{
        \textbf{Close-view generation results from \algoNameFull}.
        We generate images at continuous viewpoints with an offset of 10 degrees.
    }
    \vspace{\figmargin}
    \label{app-fig:close-view-spad}
\end{figure*}

\heading{Close viewpoints generations from \algoNameFull.}
In \cref{app-fig:close-view-spad}, we put text-conditioned multi-view generations from \algoNameFull where we increment the azimuth angle by 10 degrees per view.
We find that \algoNameFull can synthesize continuous moving views well, without content copying issues.

\subsection{User Study comparing \algoNameFull with MVDream}\label{app-sec:user-study}
We conducted a user study on the visual quality, 3D consistency, and text alignment of multi-view generations.
We distributed our questions via Amazon Mechanical Turk, where participants were given 4-view generations of \algoNameFull and MVDream~\cite{MVDream}, and asked to choose the better one satisfying the above properties. 
We found that \algoNameFull is preferred over MVDream with 59\% vs 41\%. 

\heading{Exact Instructions:} You are shown a text promt and two sets of images corresponding to 4 different views of the same object.  The views is front, left, right and back. Your task is to choose which of the sets of views is better, based on (1) consistency between different views (e.g it should represent the same object, have the same structure and colors) (2) looks better visually, (3) describes what is written in the text accurately, either Option A or Option B. 

\subsection{Training with Stable Diffusion v2.1 Weight}\label{app-sec:v21-weights}

The \algoNameFull model we evaluated in the main paper and \cref{app-sec:more-spad-results} is initialized from the weight of Stable Diffusion (SD) v1.5.
Here, we train another model initializing from the weight of the stronger SD v2.1 release.
\cref{app-fig:supp-ours-cfg-sd21} presents the multi-view generation results of this model.
Indeed, we observe better alignment with the text input, especially with longer and more complicated prompts.

This is also verified by the quantitative result.
\algoNameFull with SD v1.5 achieves a CLIP-score of \pms{29.87}{3.33}.
\algoNameFull with SD v2.1 achieves a better CLIP-score of \pms{30.39}{3.30}, which is also higher than MVDream~\cite{MVDream} initialized from the same SD v2.1 weight (\pms{30.22}{3.83}).

\subsection{Classifier-free Guidance}

Classifier-free diffusion guidance~\cite{ho2022classifierfree} is a technique used to balance the quality and diversity of images produced by diffusion models.
This method is particularly effective in class-conditional and text-conditional image generation, enhancing both the visual quality of images and their alignment with given conditions.
Inspired by~\cite{brooks2023instructpix2pix} we explore the integration of classifier-free guidance with Epipolar Attention and Plücker Embedding.
Implementing classifier-free guidance involves simultaneous training of the diffusion model for both conditional and unconditional denoising tasks.
During inference, these models' score estimates are merged. We have four different types of conditioning injected into our system:

\begin{itemize}
    \item Text ($c_T$): Injected from CLIP text-encoder similar to Vanilla Stable Diffusion. 
    \item Camera ($c_C$): Injected with timestep via Residual blocks. 
    \item Epipolar Attention ($c_E$): Injected by applying mask during self-attention. 
    \item Plücker Embedding ($c_P$): Injected by concatenation during self-attention. 
\end{itemize}

During training, we extend classifier-free guidance over all these conditions.
Therefore, our modified score estimate during inference is as follows:
\begin{align*}
    \tilde{e_{\theta}}(z_t, c_T, c_C, c_E, c_P) &= e_{\theta}(z_t, \varnothing, \varnothing, \varnothing, \varnothing)\\
    + s_T \cdot (e_{\theta}(z_t, c_T, \varnothing, \varnothing, \varnothing) &- e_{\theta}(z_t, \varnothing, \varnothing, \varnothing, \varnothing)) 
    \\
    + s_C \cdot (e_{\theta}(z_t, c_T, c_C, \varnothing, \varnothing) &- e_{\theta}(z_t, c_T, \varnothing, \varnothing, \varnothing))
    \\
    + s_E \cdot (e_{\theta}(z_t, c_T, c_C, c_E, \varnothing) &- e_{\theta}(z_t, c_T, c_C, \varnothing, \varnothing))
    \\
    + s_P \cdot (e_{\theta}(z_t, c_T, c_C, c_E, c_P) &- e_{\theta}(z_t, c_T, c_C, c_E, \varnothing))
    \label{eq:cfg2}
\end{align*}

\begin{figure*}[t]
    \vspace{\pagetopmargin}
    \hspace{-0.03\linewidth}
    \includegraphics[width=1.06\linewidth]{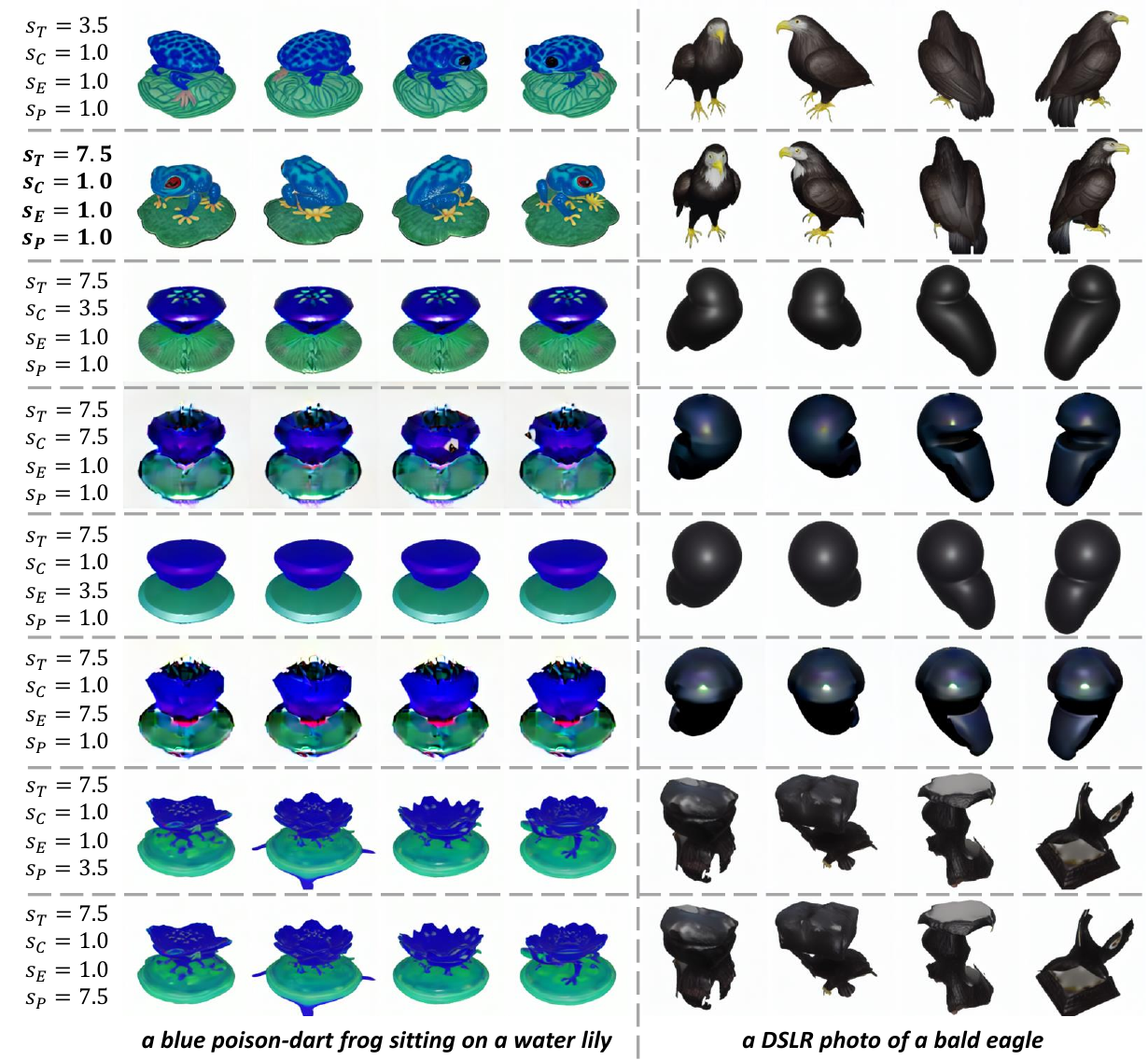}
    \vspace{\figcapmargin}
    \vspace{-3mm}
    \caption{
        \textbf{Ablation study regarding the classifier-free guidance scales.}
        Using a large scale of $s_T = 7.5$ for text conditioning works the best (row 2), while increasing scales for camera embedding, Epipolar Attention, and Plücker Embedding all leads to over-saturated images.
    }
    \vspace{\figmargin}
    \label{app-fig:cfg-ablation}
\end{figure*}

\heading{Outcome:} As shown in \cref{app-fig:cfg-ablation}, we find that classifier-free guidance beyond text conditioning does not provide additional benefits, and rather leads to over-saturated generations.
This also aligns with our observations on MVDream.

\subsection{Joint Multi-View Inference}\label{sec:mv-inference}

Concurrent multi-view diffusion models~\cite{MVDream,SyncDreamer} are limited to generating the same number of views they were trained on during testing.
However, generating a high-quality 3D asset by e.g. training a NeRF model usually requires more than ten views of the asset.
A naive solution is to use more views during training, which leads to quadratically increasing training costs due to the use of 3D self-attention. Instead, we propose a joint multi-view inference technique, which enables generating an infinite number of views using a model trained with fewer views.

Assume that we want to generate $M$ views with a two-view model.
We first initialize $M$ noise maps $\{\img_T^i\}_{i=1}^M$, and then iteratively denoise all possible pairs of views:
\begin{equation}\label{eq:pairwise-denoise}
    (\img_{t-1}^i, \img_{t-1}^j) = \mathrm{Denoise}(\img_{t}^i, \img_{t}^j, \dmmodel),\ \forall\ i, j \in [1, M], i \ne j.
\end{equation}
Since the model is only trained on both views with the same noise level (\ie timestep $t$), we sample $(i, j)$ pairs without replacement and make sure to go over all possible combinations uniformly via simple heuristics.

\heading{Outcome:} We find that this experiment trades off 3D consistency, as it only allows cross-view communication between two views at any given timestep of generation.

\subsection{Fréchet Inception Distance (FID) Results}

Compared to Vanilla MV-DM with an FID score of 55.25, our full model \algoNameFull achieves a better FID score of 52.77 which shows further evidence of improvement in 2D generation quality.   

\heading{FID Comparison with MVDream.} Since our model generates images at random views, it has a much larger pose distribution mismatch compared to MVDream which uses orthogonal (90-degree varying) views in both ground-truth and generated images. Due to this reason, our FID cannot be compared directly with MVDream (trained with v2.1) which is reported to be $32.06$ in the original work.

\clearpage
\begin{figure*}[t]
    \vspace{\pagetopmargin}
    \centering
    \includegraphics[width=1.0\linewidth]{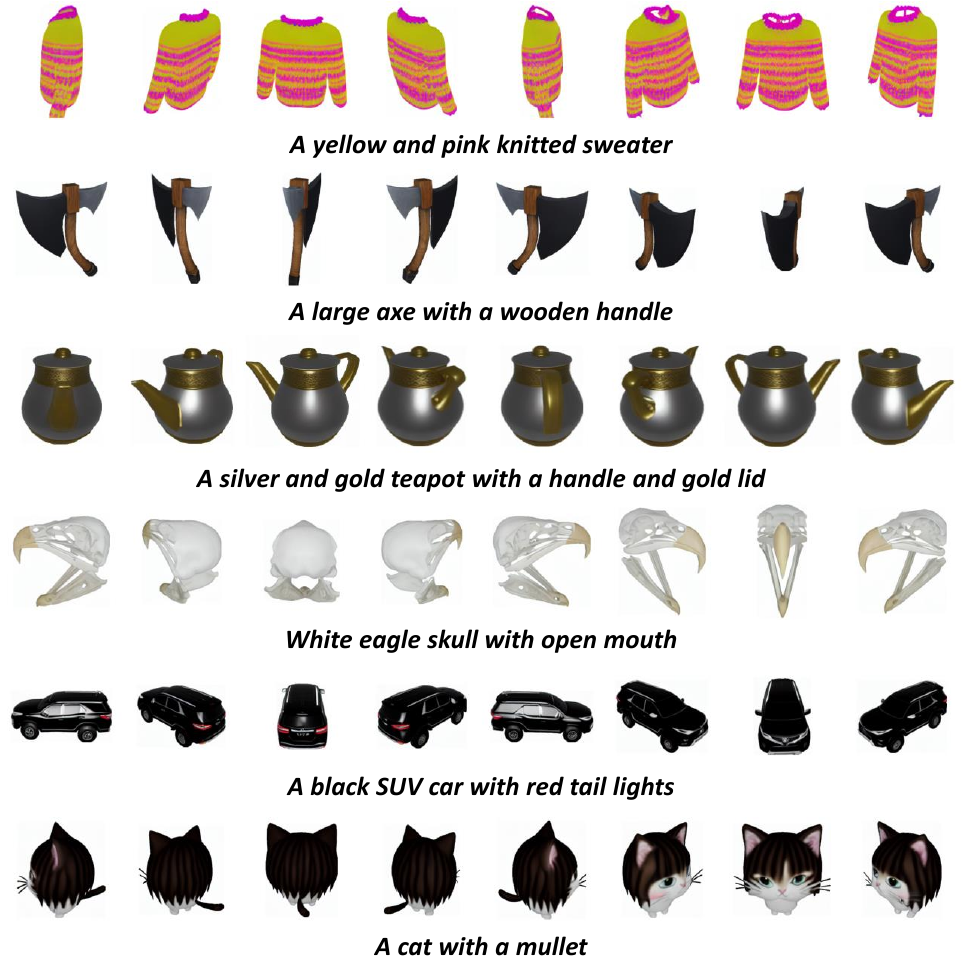}
    \vspace{\figcapmargin}
    \vspace{-3mm}
    \caption{
        \textbf{More multi-view generation results with \algoNameFull.}
        The tested model is initialized with the weight of Stable Diffusion v1.5, and fine-tuned on Objaverse rendered images (same as Fig.~{1} in the main paper).
    }
    \vspace{\figmargin}
    \label{app-fig:supp-ours-cfg}
\end{figure*}

\clearpage
\begin{figure*}[t]
    \vspace{\pagetopmargin}
    \vspace{-5mm}
    \centering
    \includegraphics[width=1.0\linewidth]{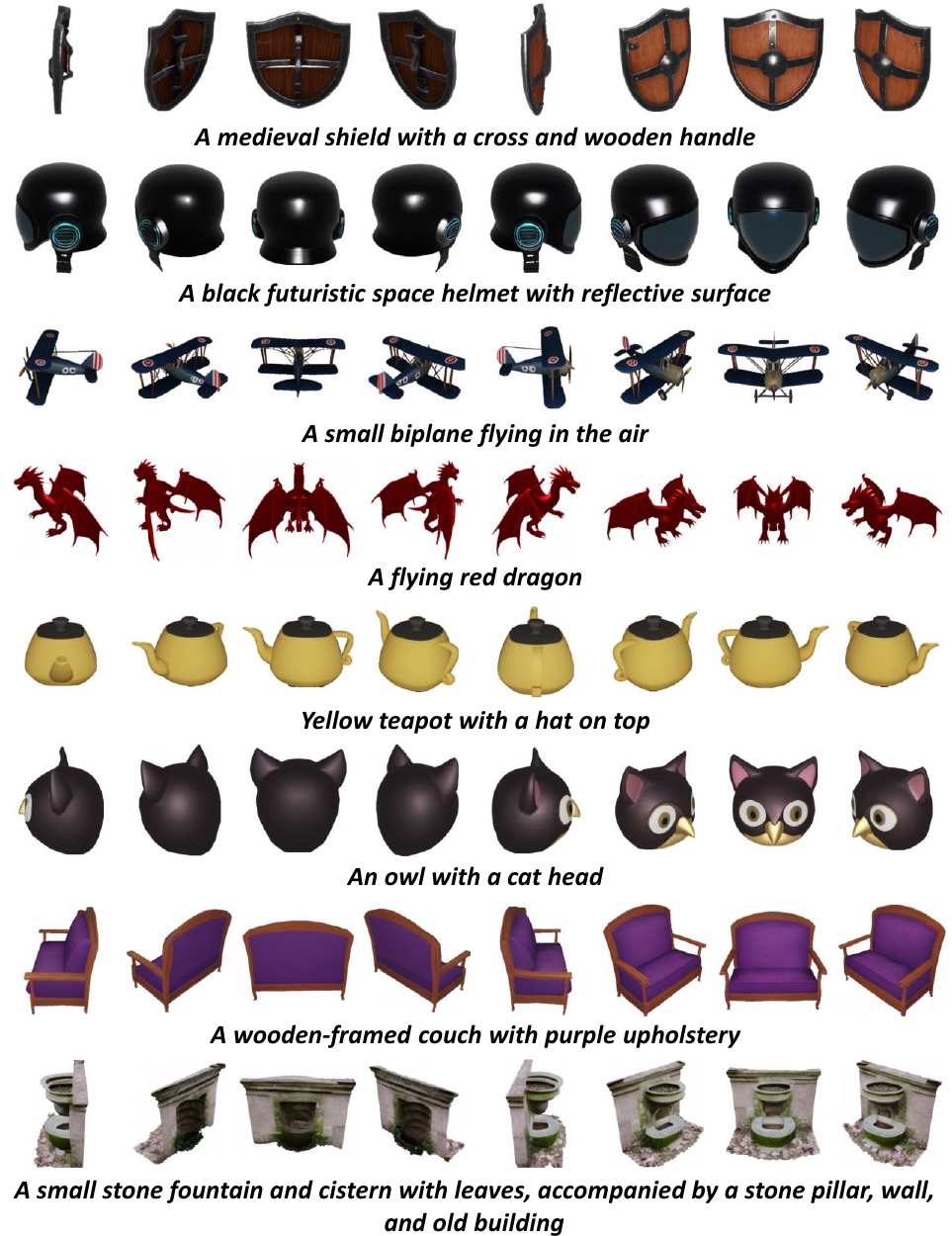}
    \vspace{\figcapmargin}
    \vspace{-3mm}
    \caption{
        \textbf{More multi-view generation results with \algoNameFull.}
        The tested model is initialized with the weight of Stable Diffusion v2.1, and fine-tuned on Objaverse rendered images.
        Compared to results in \cref{app-fig:supp-ours-cfg} which adopts the weight of Stable Diffusion v1.5, this model is able to follow more complicated text prompts.
    }
    \vspace{\figmargin}
    \label{app-fig:supp-ours-cfg-sd21}
\end{figure*}

\clearpage
